\documentclass[sigconf,natbib, nonacm, authorversion]{acmart}
\AtBeginDocument{%
  }

\setcopyright{acmlicensed}
\copyrightyear{2025}
\acmYear{2025}
\acmDOI{XXXXXXX.XXXXXXX}

\usepackage{multirow}
\usepackage{tabularx}

\begin{document}

\title{Reading Between the Timelines: RAG for Answering Diachronic Questions}

\author{Kwun Hang Lau}
\affiliation{%
  \institution{Hong Kong University of Science and Technology}
  \city{Hong Kong}
  \country{China}
}

\author{Ruiyuan	Zhang}
\affiliation{%
  \institution{Hong Kong University of Science and Technology}
  \city{Hong Kong}
  \country{China}
}

\author{Weijie Shi}
\affiliation{%
  \institution{Hong Kong University of Science and Technology}
  \city{Hong Kong}
  \country{China}
}

\author{Xiaofang Zhou}
\affiliation{%
  \institution{Hong Kong University of Science and Technology}
  \city{Hong Kong}
  \country{China}
}

\author{Xiaojun	Cheng}
\affiliation{%
  \institution{China Unicom (Hong Kong) Operation Ltd}
  \city{Hong Kong}
  \country{China}
}

\begin{abstract}
While Retrieval-Augmented Generation (RAG) excels at injecting static, factual knowledge into Large Language Models (LLMs), it exhibits a critical deficit in handling longitudinal queries that require tracking entities and phenomena across time. This blind spot arises because conventional, semantically-driven retrieval methods are not equipped to gather evidence that is both topically relevant and temporally coherent for a specified duration. We address this challenge by proposing a new framework that fundamentally redesigns the RAG pipeline to infuse temporal logic. Our methodology begins by disentangling a user's query into its core subject and its temporal window. It then employs a specialized retriever that calibrates semantic matching against temporal relevance, ensuring the collection of a contiguous evidence set that spans the entire queried period. To enable rigorous evaluation of this capability, we also introduce the Analytical Diachronic Question Answering Benchmark (ADQAB), a challenging evaluation suite grounded in a hybrid corpus of real and synthetic financial news. Empirical results on ADQAB show that our approach yields substantial gains in answer accuracy, surpassing standard RAG implementations by 13\% to 27\%. This work provides a validated pathway toward RAG systems capable of performing the nuanced, evolutionary analysis required for complex, real-world questions. The dataset and code for this study are publicly available at \url{https://github.com/kwunhang/TA-RAG}.
\end{abstract}

\maketitle

\section{Introduction}
Retrieval‑Augmented Generation (RAG) has been shown to enhance Large Language Models (LLMs) by injecting external evidence at inference time, thereby improving factual accuracy and mitigating hallucinations \cite{lewis2020retrieval, gao2024retrievalaugmented}. This mechanism broadens the practical scope of LLMs for knowledge‑intensive tasks \cite{lewis2021retrievalaugmentedgenerationknowledgeintensivenlp, ovadia2024finetuningretrievalcomparingknowledge, app15063134, 2023paperqaretrievalaugmentedgenerativeagent}.

Time-sensitive queries, such as “What was Apple’s stock price on March 1st, 2025?” , present a highly demanding yet challenging task for Retrieval-Augmented Generation (RAG) systems. Recent studies have begun to tackle this issue by improving temporal awareness in retrieval processes \cite{gade2024itsabouttime} \cite{wu2024timesensitive}. However, RAG systems still struggle with an even more complex category of questions known as analytical diachronic questions (ADQ) —queries that require synthesizing and analyzing information across extended time periods.

A typical ADQ could be phrased as: “Summarize the trend in Apple’s stock price from 2015 to 2025.” Such queries involve three essential components: a specific entity (e.g., Apple), a broad time range rather than a single point in time, and a summarization or analytical task . Unlike simpler factual questions, the answer to an ADQ cannot be found in a single document or text chunk. Instead, it requires comprehensive retrieval of relevant information across a large corpus covering the entire specified time span, followed by coherent synthesis of that information. A key limitation of current RAG systems in handling ADQs is their inconsistent ability to retrieve evidence that fully covers the required time period. Recent efforts have attempted to address this challenge. For instance, TS-Retriever \cite{wu2024timesensitive} improves temporal relevance through supervised contrastive learning, while TempRALM \cite{gade2024itsabouttime} integrates temporal filtering mechanisms without requiring model retraining. Despite these advances, ADQs demand the aggregation of multiple temporally distributed sources into a unified, coherent narrative—a requirement that current methods are not specifically designed to fulfill.

This challenge differs significantly from traditional multi-hop reasoning tasks, which typically involve navigating between different entities connected by explicit relationships \cite{jia2018tempquestions, chen2021timeqa}. In contrast, ADQs require temporal coherence across retrieved documents, ensuring that the synthesized output reflects a continuous and accurate representation of how a particular entity or phenomenon evolved over time. Current approaches, which often focus on isolated facts or direct temporal links between entities, fall short in addressing this need.

Furthermore, there is a notable lack of dedicated benchmark datasets that evaluate a system’s ability to perform multi-step temporal reasoning over extended timeframes. This gap hinders progress toward building RAG systems capable of answering complex, real-world analytical questions that require both deep temporal understanding and contextual synthesis.

To address these limitations, we propose Time-Aware RAG (TA-RAG), a retrieval-generation framework designed for queries spanning long time horizons. Our approach integrates temporal awareness at every pipeline stage, specifically targeting temporally grounded scenarios. TA-RAG first parses temporal expressions in queries, splitting them into semantic and temporal components. The Time-Sensitive Retriever then uses document event intervals and temporal query embeddings to retrieve documents that are both semantically relevant and temporally consistent. Finally, the LLM generates answers with structured temporal context, ensuring factual accuracy and temporal coherence. Additionally, we introduce ADQAB, a benchmark for diachronic questions requiring broad temporal retrieval and cross-time synthesis.

The main contributions of this work are summarized as follows:\begin{itemize}

    \item We propose the Time-Aware RAG framework, which improves temporal sensitivity in retrieval, ensures broad coverage of relevant time spans, and supports advanced reasoning for diachronic questions.
    \item We present a new benchmark dataset designed to evaluate RAG systems' ability to answer complex, time-based questions that require synthesizing information across multiple time periods.
    \item Through empirical evaluation, we demonstrate the effectiveness of our proposed framework and benchmark in advancing the state-of-the-art for answering complex questions across time periods.
\end{itemize}
\section{Related Work}
\subsection{Temporal Retrieval Augmented Generation}
RAG has emerged as a powerful paradigm for enhancing the capabilities of LLMs by grounding them in external knowledge sources~\cite{lewis2020retrieval}. A growing body of work has focused on improving various aspects of RAG, including retrieval mechanisms, embedding models, and generation strategies~\cite{gao2024retrievalaugmented}. However, the effectiveness of standard RAG is significantly diminished when applied to temporal queries, i.e., questions that require understanding and reasoning about time. Recent research has shown that current retrievers exhibit limitations in handling time-sensitive queries~\cite{wu2024timesensitive}. Existing RAG approaches often struggle in the temporal domain primarily because standard retrieval mechanisms predominantly emphasize semantic similarity, overlooking the critical temporal constraints embedded within many queries. This can result in the retrieval of outdated or temporally irrelevant information, leading to inaccurate or unreliable generated responses. For example, a query about "the unemployment rate in 2010s" necessitates retrievers to understand not only the semantics of "unemployment rate" but also to accurately capture the temporal constraint "in 2010s". Current embedding-based retrievers, while effective at capturing semantic similarity, often fail to adequately prioritize temporal relevance, leading to the retrieval of documents that may be semantically related but temporally inappropriate.

Recent research has begun to address these challenges, albeit with limitations. TempRALM \cite{gade2024itsabouttime} proposed a Retrieval Augmented Language Model that considers both semantic and temporal relevance during document selection. MRAG \cite{siyue2024mrag} introduced a modular retrieval framework that aims to enhance time-sensitive question answering with LLMs. The framework focuses on time-sensitive questions, and it does not include a model to process the implicit time constraints. FAITH \cite{jia24Faithful} introduced a question-answering system that operates over heterogeneous sources and aims for faithful answering by enforcing temporal constraints. A key contribution of FAITH is its method for transforming implicit temporal constraints into explicit ones by recursively invoking the QA system. It mainly focuses on the faithfulness of the answering in temporal QA. BiTimeBERT \cite{2023bitimebert, 2025bitimebert2_0} introduce new pretraining tasks, to improve the model's performance on time-related tasks . TS-Retriever \cite{wu2024timesensitive} explore the time-sensitive queries in RAG models and propose their Time-Sensitive Retriever, adopting contrastive learning with tailored negative sample pairs for temporal constraints to train the retriever. TempRetriever proposes a fusion-based retrieval that incorporates time information into the retrieval process, enhancing performance in time-aware passage retrieval \cite{abdallah2025tempretrieverfusionbasedtemporaldense}. Despite these advancements, there remains significant room for improvement, particularly in handling more complex temporal reasoning scenarios, such as those involving intricate temporal relationships or implicit time constraints.

\subsection{Temporal Question Answering}
Temporal Question Answering (TQA) is a foundational area for our research, focusing on systems that comprehend and reason about temporal information in questions. Success in TQA hinges on two core capabilities: (1) Temporal Relation Understanding, which involves interpreting temporal expressions (e.g., dates, times, durations) and their relationships (e.g., before, after, during) \cite{wei2023menatqa,zhang2024TECLongBench, chu-etal-2024-timebench, su-etal-2024-living,10.1145/3589334.3645376}, and (2) Time-Aware Knowledge Retrieval, the ability to access time-sensitive facts.

A primary challenge in TQA is that Large Language Models (LLMs) are often trained on static datasets, making their knowledge outdated. Consequently, they struggle to answer questions about recent events that fall beyond their training data's knowledge cutoff. While continuous pre-training can inject new knowledge, this approach is computationally demanding and risks catastrophic forgetting \cite{Dhingra2022LLMasTKB}. A more prevalent and flexible solution is to augment LLMs with external, up-to-date knowledge bases at inference time, a paradigm central to our work.

To benchmark progress, numerous TQA datasets have been developed. Early influential work includes TempQuestions \cite{jia2018tempquestions}, which contains questions with both explicit and implicit temporal elements. Subsequent datasets address more complex scenarios. For instance, MULTITQ \cite{chen2023MULTITQ} and related works \cite{qian2024timer4, liu2023LGQA,chen-etal-2024-temporal-ARI} tackle challenges of mixed temporal granularities over knowledge graphs. TimeQA \cite{chen2021timeqa} specifically targets questions requiring reasoning over facts that change over time. For long-range historical text, ChroniclingAmericaQA \cite{piryani2021chroniclingamericaqa} provides a large-scale dataset from digitized newspaper archives. To evaluate a model's ability to adapt to new information, StreamingQA \cite{liška2022streamingqa} presents questions based on a stream of timestamped news articles. More recently, datasets have focused on specialized temporal challenges. TIQ \cite{jia2024tiq} emphasizes questions with implicit temporal constraints. TS-Retriever \cite{wu2024timesensitive} created a dataset to evaluate time-sensitive Retrieval-Augmented Generation (RAG), and FinTMMBench \cite{zhu2025fintmmbenchbenchmarkingtemporalawaremultimodal} introduces a multi-modal corpus for temporal reasoning in the financial domain. This progression of increasingly complex and diverse TQA datasets underscores the critical need for advanced, temporally-aware retrieval mechanisms, which motivates our proposed framework.

\begin{table}
\centering
\caption{Summary of notations}
\vspace{-3mm}
\label{tab:notation}
{
\begin{tabular}{|r|p{0.6\columnwidth}|}
\hline
{Notation} & {Definition} \\
\hline \hline
$Q_s$ & Semantic content of input query \\
\hline
$Q_t$ & explicit or implicit temporal constraints  of input query\\
\hline
$D$ & Document corpus\\
\hline
$D'$ & The retrieval document set, where $D' \subset D$\\
\hline
$t$ &  A time-point\\
\hline
$T$ &  A time-interval\\
\hline
$\hat{t}_{pub}$ &   The estimated document’s publication time \\
\hline
$T_{e}$ / $[t_{e,start}, t_{e,end})$ &  The corresponding time interval of event\\
\hline
$\{T_{q}\}$ &  A set of query time intervals, capturing the specified temporal constraints of the query Q\\
\hline
$q_{core}$ &  Temporally neutral core query , representing the main semantic topic of the query Q\\
\hline
$\text{synthesize}(q_{core}, t_i)$ & Synthesize query by combining $q_{qcore}$ and $t_i$ \\
\hline
$e_{hypo}$ &  hypothetical temporal query embedding construct from $\text{synthesize}(q_{core}, t_i)$ \\
\hline
$C_{real}$ & Real News Corpus from FNSPID \\
\hline
$ C_{synth}$ & Synthetics News Corpus \\
\hline
$C_{final}$ & Final Augmented Corpus \\
\hline
\end{tabular}}
\vspace{-5mm}
\end{table}

\section{Preliminary}
\label{sec:preliminary}
\subsection{Problem Formulation}
\label{ssec:problem_formulation}
We begin by formally defining the problem. Table 1 summarizes the key notations used in this paper. Although RAG has markedly improved open‑domain question answering, conventional pipelines still struggle with queries that require reasoning over temporal dynamics. We focus on a particularly demanding subset—\textbf{analytical diachronic questions (ADQ)}—that call for analyzing trends, summarizing evolution, detecting change, or performing cross‑period comparisons anchored to explicit time spans or reference points. Solving such tasks entails more than locating a single, time‑stamped fact; it requires aggregating and synthesizing evidence drawn from multiple moments within the query’s temporal window and integrating those pieces into a coherent, time‑aware answer.

Formally, let $Q$ be an input query comprising semantic content $Q_s$ and explicit or implicit temporal constraints $Q_t$. These constraints often define a time interval $[t_{start}, t_{end}]$, a set of discrete time points $\{t_1, t_2, \dots\}$, or relative temporal conditions (e.g., ``before $t_x$'', ``after $t_y$''). Let $D = \{d_1, d_2, \dots, d_n\}$ be a corpus where documents $d_i$ possess associated \textbf{temporal information}, representing the time period(s) or point(s) to which the document's content is relevant. The core challenge in ADQ answering within a RAG framework involves addressing key difficulties in both the retrieval and generation stages.

\subsection{Problem Analysis}
\label{ssec:problem_analysis}
Our primary focus in this work is on the intricate challenges encountered during the retrieval stage. The objective here is to identify a document subset $D'\subset D$ that is not only semantically relevant to $Q_s$ but also temporally congruent with $Q_t$. Existing retrieval mechanisms, frequently optimized to maximize semantic similarity, encounter several critical obstacles when faced with the nuanced demands of diachronic queries.

First, a pervasive issue is the \textbf{Neglect of Temporal Constraints}. Retrieval models that predominantly rely on vector similarity or keyword matching for $Q_s$ can readily identify documents that are semantically aligned but whose temporal metadata are inconsistent with the query's specified timeframe $Q_t$\cite{wu2024timesensitive}. For instance, information pertinent to a future decade might be retrieved for a query concerning the past one, leading to anachronistic and incorrect answers. Second, ensuring adequate \textbf{Temporal Coverage} across the full extent of $Q_t$ is paramount, yet particularly challenging. For diachronic questions spanning a considerable interval (e.g., $[t_{start}, t_{end}]$), a robust retrieval strategy must furnish documents that reflect the state of affairs not merely at the interval's endpoints but also at various representative nodes within the span. Such comprehensive coverage is essential for tasks such as analyzing evolutionary trends or constructing a holistic summary of developments over the specified period. Without it, the synthesized answer may be based on an incomplete or biased view of the temporal landscape.

Moreover, retriever can exhibit \textbf{Temporal Endpoint Bias}, disproportionately favoring documents that explicitly mention $t_{start}$ or $t_{end}$ \cite{siyue2024mrag}, thus failing to provide a representative set of evidence covering the entire duration. This "temporal endpoint bias" arises because documents that explicitly mention the start or end of the query range may receive higher surface relevance scores through standard semantic searchers, especially if these boundary markers are part of the query string, thereby overlooking documents relevant to the intervening periods. An effective diachronic-aware retriever must overcome both the neglect of constraints and the bias issues to ensure comprehensive temporal coverage.

Existing benchmarks for temporal question answering often do not fully address the combined challenges of ensuring broad and representative temporal coverage in the retrieval phase while also demanding accurate and nuanced diachronic responses in the generation phase. Recognizing these gaps, this paper introduces novel RAG techniques and corresponding evaluation methods specifically engineered to address the rigorous requirements of ADQ, with a particular emphasis on overcoming these critical retrieval challenges.
\section{Methodology: TA-RAG}
\label{TA-RAG}
\begin{figure*}[t!]
  \centering
  \includegraphics[width=\textwidth]{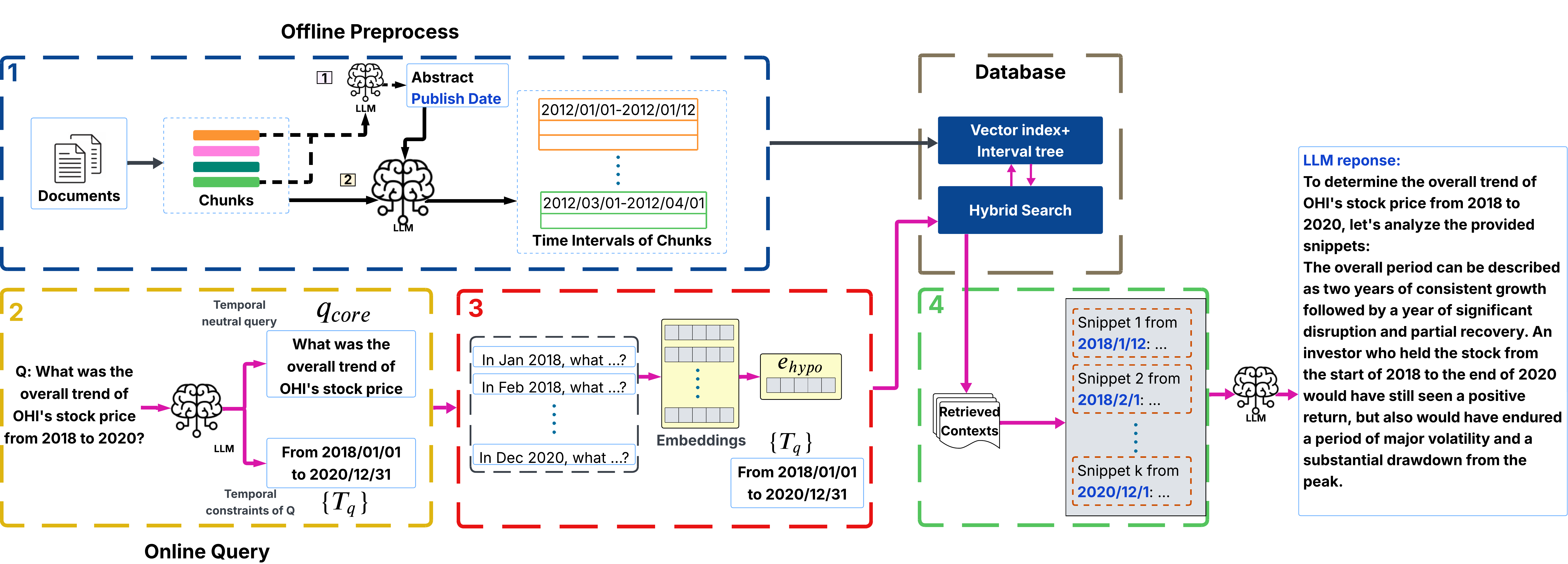}
  \caption{An overview of the TA-RAG framework, consisting of four key modules: Time Information Extraction, Question Processing, Time-filtering Retrieval Strategy and Temporal Context Structuring for Generation.}
  \label{fig:TA-RAG}
\end{figure*}

RAG systems frequently deliver suboptimal results for time-sensitive questions due to the inherent difficulty standard retrievers face in interpreting temporal constraints alongside semantic relevance. To overcome this, we proposed the Time-aware Retrieval-Augmented Generation (TA-RAG) framework (as shown in figure\ref{fig:TA-RAG}). The following subsections present the four main components: (1) extracting temporal information from source documents, (2) processing time-sensitive queries, (3) performing temporally-aware retrieval, and (4) structuring context for generation. This holistic approach aims to enhance the accuracy of retrieved evidence and the temporal reasoning capabilities of the final generation model.
\subsection{Time Information Extraction}
\label{ssec:Time Information Extraction}
A fundamental element of TA-RAG is the comprehensive preprocessing of source documents to extract and normalize all temporal information,  irrespective of their original granularity, into a consistent time-interval representation. Our empirical analysis of diverse textual sources, such as news articles and annual reports, reveals a common pattern: crucial document-level temporal context, including publication dates and the primary temporal scope of events, is frequently concentrated in the initial and terminal sections of documents. Moreover, we observe that within the body of longer texts, subsequent references to dates often become abbreviated (e.g., mentioning only the day/weekday/month after a full date's initial introduction), risking the loss of complete temporal anchors when documents are segmented into smaller, isolated chunks. Thus, standard chunking techniques can obscure these vital temporal cues often found at document extremities. To mitigate this information loss and enhance the precision of temporal metadata extraction, we employ a two-stage temporal annotation process using LLM.  This approach is designed to mitigate information loss and enhance the precision of temporal metadata extraction. 

In the first stage, the LLM analyzes the initial and final segments of each document. From these segments, it estimates an overall document publication time($\hat{t}_{pub}$) and generates a brief (1-2 sentence) abstract. These outputs serve as crucial contextual input for the subsequent stage, refining its ability to annotate finer-grained temporal details.
In the second stage, leveraging the document-level context established previously, the LLM processes each individual chunk. For each chunk, it identifies contained events and extracts their corresponding temporal intervals, denoted as $T_e = [t_{e,start}, t_{e,end})$. 

The final indexed representation for each chunk thus includes its textual content, its vector embedding, and this extracted set of event-specific time intervals $\{T_e\}$, which directly facilitates our time-filtered retrieval mechanism.
\vspace{-0.3cm}
\subsection{Question Processing}
\label{ssec:Question Processing}
Upon receiving a time-sensitive question $Q$, our system initiates a question processing phase. The central aim is to effectively disentangle the core semantic intent of the question from its associated temporal specifications. To achieve this decomposition, we leverage a LLM to decompose the question $Q$ into two key outputs: a temporally neutral core query, $q_{core}$, representing the main semantic topic, and a set of query time intervals, $\{T_q\}$, capturing all temporal constraints (e.g., $T_q$ representing "before 2015"). This decomposition enables a nuanced approach where semantic relevance and temporal compliance can be evaluated distinctly yet cohesively. Our strategy for partitioning semantic and temporal aspects of the query aligns conceptually with the methodology presented in MRAG \cite{siyue2024mrag}.

\subsection{Temporally-Aware Retrieval Strategy}
\label{ssec:Time-filtering Retrieval}
To retrieve document chunks that are both semantically relevant and temporally consistent with a given query $Q$, our approach enhances the query representation itself before employing a filter search. Standard embeddings of the core query, $q_{core}$, may not adequately capture specific temporal information needs. We therefore construct a hypothetical temporal query embedding, $e_{hypo}$, designed to be sensitive to the query's specified time constraints, $\{T_q\}$.
The construction of $e_{hypo}$ begins by sampling the temporal landscape defined by $\{T_q\}$. We generate a set of $n$ discrete temporal anchor points, $\{t_i\}_{i=1}^n$, which collectively span the interval(s) in $\{T_q\}$. The granularity of these anchor points is crucial and determined dynamically based on the breadth of the query's temporal range to ensure representative coverage and manageable computational cost. Our heuristic is to set the granularity one level finer than the primary unit of the query's time range.
For example, given a query that spans several decades, we sample at the year level. For a query in several months to a few years, we would sample at the month level. This adaptive granularity ensures that the subsequent retrieval appropriately captures the temporal nuances.

For each temporal anchor point $t_i$, we synthesize a query variant by prepending a natural language phrase representing $t_i$ to the core query $q_{core}$ (e.g. "In January 2011, [core query]"). The embeddings of these $n$ variants are then averaged to produce the final query embedding:
$$e_{hypo} = \frac{1}{n} \sum_{i=1}^{n} \text{Embed}(\text{synthesize}(q_{core}, t_i))$$
where $\text{Embed}(\cdot)$ is the text embedding function and $\text{synthesize}(\cdot)$ denotes the creation of query variant. This averaging yields a robust embedding that reflects the query's semantic intent across the entire specified temporal scope.

Retrieval then proceeds in two steps. First, a temporal filter is applied to the document chunk corpus. This filter selects a candidate set, containing only those chunks whose associated event time intervals, $T_e$, overlap with the query's temporal specifications $T_q$ (i.e. , $T_q \cap T_e \neq \emptyset$). In the second stage, these temporally-relevant candidate chunks are ranked by their semantic similarity to the $e_{hypo}$ embedding. The top-$k$ chunks from this ranking are then returned, ensuring both temporal consistency and strong semantic relevance to the user's query. To implement this two-stage retrieval efficiently, a static interval-tree is built on the chunks' time intervals for fast temporal filtering, followed by a search on a flat index of embeddings for the temporally relevant candidates. In this work, we focus on corpus whose documents contain explicit or inferable temporal cues.

\subsection{Temporal Context Structuring for Generation}
\label{Temporal Context Structuring for Generation}
Following the retrieval of temporally relevant documents, the final step is to structure the information coherently to facilitate synthesis by the downstream LLM. We contend that the mere aggregation of relevant chunks is insufficient; the temporal ordering and explicit signaling of time are paramount for the LLM to construct a temporally sound and factually accurate narrative.

Our approach incorporates a temporal context structuring phase. The core operation involves organizing the retrieved set of documents in chronological order based on their estimated publication time, $\hat{t}_{pub}$, which were previously determined in Section \ref{ssec:Time Information Extraction}. This ordered sequence of chunks, where each chunk is explicitly accompanied by its $\hat{t}_{pub}$ metadata, is then concatenated to form the final contextual input for the LLM generator.

This structured input enables the model to better understand the evolution of events or information over time, leading to the generation of answers that are not only semantically relevant but also improved temporal consistency. This step is crucial for addressing complex queries where temporal reasoning is key to answering correctly.

\section{ADQAB: Analytical Diachronic Question Answering Benchmark}
\label{sec:dataset}
\begin{figure*}[t!]
  \centering
  \includegraphics[width=0.9\textwidth]{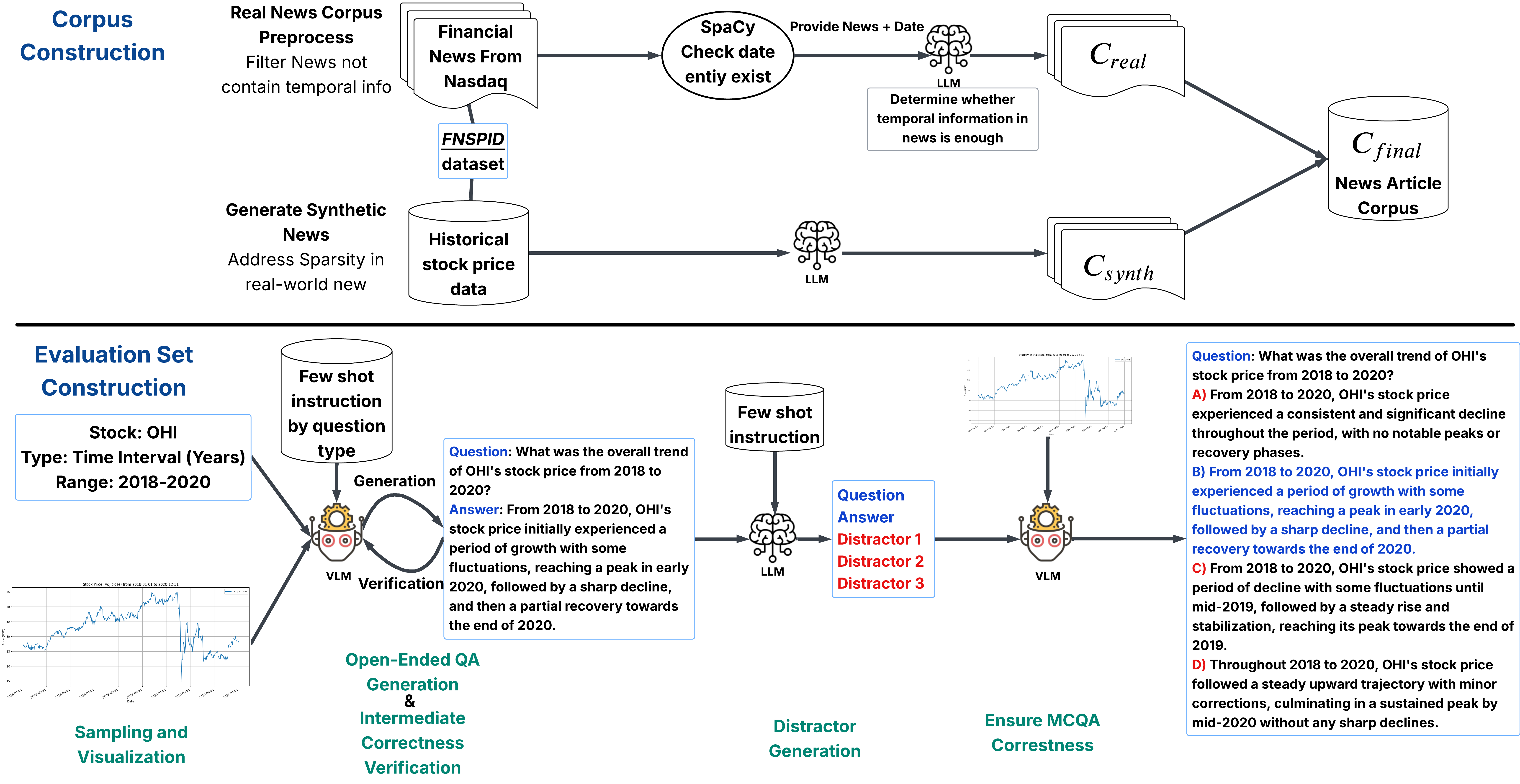}
  \caption{Design of the Analytical Diachronic Question Answering Benchmark.}
  \label{fig:ADQAB}
\end{figure*}
To rigorously evaluate the capacity of RAG systems to address the complexities of ADQ, as formulated in Section \ref{sec:preliminary}, we introduce \textbf{ADQAB}, a novel benchmark dataset. ADQAB comprises two core components: a comprehensive \textbf{Corpus} designed to provide broad temporal coverage, and a challenging \textbf{Evaluation Set} consisting of multiple-choice questions that require temporal reasoning across different time periods. The construction methodology integrates real-world financial news with targeted synthetic data generation to ensure both authenticity and knowledge continuity, shown in Figure\ref{fig:ADQAB}. 
\subsection{Corpus Construction}
\label{ssec:corpus_construction}
The construction of the corpus aimed to create a temporally dense collection of documents suitable for answering questions spanning extended periods. This involved curating relevant real-world news and systematically augmenting it with synthetic data to mitigate temporal sparsity.

\textbf{Real News Corpus Curation and Filtering.}
We initiated the process by using the Financial News and Social Perception Index Dataset (FNSPID) \cite{dong2024fnspid}, which contains Nasdaq-supplied news articles from 2012 to 2022. Our scope was focused on articles pertaining to 25 selected publicly traded stocks within this time frame. Recognizing the critical need for reliable temporal grounding, we implemented a stringent two-stage filtering protocol.
\begin{enumerate}
    \item \textbf{Explicit Date Filtering:} We first utilized SpaCy's Named Entity Recognition (NER) capabilities to identify and retain only those articles containing explicit date mentions that included a year specification \cite{spacy2}. This step ensures a baseline level of temporal anchoring.
    \item \textbf{Internal Temporal Context Verification:} To ascertain richer temporal relevance beyond the publication date, we employed a LLM\footnote{The model used for internal temporal context verification was LLaMA-3.1-7B-Instruct\cite{llama3modelcard}.}. Each candidate article, along with its publication date metadata, was evaluated by the LLM. The objective was to determine whether the narrative content provided sufficient internal temporal cues to allow for an approximate timeline reconstruction, independent of the explicit metadata. Articles judged by the LLM to have sufficient internal temporal context were retained.
\end{enumerate}
The resulting collection of filtered, temporally relevant real news articles constitutes the $C_{real}$ component of our corpus.

\textbf{Addressing Sparsity via Synthetic News Generation.}
Preliminary analysis of $C_{real}$ revealed potential temporal gaps, where certain stock-month combinations lacked corresponding news coverage. Such gaps could render diachronic queries unanswerable if they fall within these periods. To mitigate this issue and foster a more continuous knowledge landscape, we generated synthetic news articles ($C_{synth}$). This process utilized an LLM\footnote{The model used for synthetic news generation was LLaMA-3.3-70B-Instruct.}. For each of the 25 selected stocks, we generated one synthetic article per month for the entire period from January 2012 to December 2022. The generation was conditioned on the historical stock price data for the specific stock and month. The LLM was prompted to create plausible, concise news-like summaries reflecting market events or sentiments consistent with the observed price trends during that month. The primary goal of this synthetic generation was to enhance temporal density and improve the potential answerability of queries requiring information across diverse time points.

\textbf{Final Augmented Corpus.}
The definitive corpus employed in our experiments, denoted $C_{final}$, is the union of the curated real news and the generated synthetic articles:
$$C_{final} = C_{real} \cup C_{synth}$$
This augmented corpus provides a temporally richer and more contiguous foundation, specifically designed to support the retrieval phase of RAG systems tackling diachronic questions.

\subsection{Evaluation Set Generation}
\label{ssec:evaluation_set_generation}
To rigorously evaluate the temporal reasoning capabilities of RAG systems on ADQ tasks, we constructed a dedicated evaluation set, consisting of Multiple Choice Questions Answering(MCQA). We generate questions and a set of shuffled choices given the historical stock price,  and then verify their correctness.


\textbf{Temporal Query Types.} The questions in evaluation set are designed to probe distinct temporal reasoning patterns relevant to analyzing stock price trends. We categorize these into three fundamental types, each targeting different aspects of temporal understanding. The primary characteristics and example questions focusing on stock price trends are summarized in Table \ref{tab:query_types}.

\begin{itemize}
    \item \textbf{Specific Time Period:} These questions assess the understanding of stock price trends within a single, explicitly defined calendar year. We focus on year-level granularity for this query type when asking about trends, as discerning a clear trend from news within a single month can be challenging due to signal sparsity.
    
    \item \textbf{Before/After:} This category tests the ability to identify and characterize stock price trends over periods that precede or follow a specific temporal anchor point (which can be a year or a month). The focus is on the stock's behavior relative to this anchor.
    
    \item \textbf{Time Range:} This query type is crucial for detailed trend analysis within an explicitly defined, bounded temporal interval $[t_{start}, t_{end}]$. It includes both analyses over year-defined intervals and, importantly, fine-grained month-level trend analysis within specific multi-month windows.
\end{itemize}

\begin{table}[htbp]
\centering
\caption{ADQAB Temporal Query Types and Example Stock Price Trend Questions}
\label{tab:query_types}
\begin{tabular}{@{}p{0.22\linewidth} p{0.18\linewidth} p{0.5\linewidth}@{}}
\toprule
\textbf{Query Type} & \textbf{Granularity Focus} & \textbf{Example Question (Stock Price Trend)} \\
\midrule
Specific Time Period & Year & What was the general stock price trend for Company X during 2018? \\
\midrule
Before/After & Year (Anchor) & What was the predominant stock price trend for Company Y before the start of 2020? \\
\cmidrule{2-3}
 & Month (Anchor) & Characterize the stock price trend for Company Z in the period after June 2021. \\
\midrule
Time Interval & Years & Analyze the overall stock price trend of Company A from 2017 to 2019. \\
\cmidrule{2-3}
 & Months & Describe the stock price trend for Company B from May 2015 to January 2016. \\
\bottomrule
\end{tabular}
\end{table}

The design of these query types, particularly the inclusion of month-level trend analysis within the "Time Range" category, allows ADQAB to rigorously assess a system's capability for nuanced temporal reasoning about stock price trends.


\textbf{Generation Methodology}
We developed a novel multi-stage generation pipeline that leverages Vision-Language Models(VLMs)\footnote{The VLM used for the evaluation set generation is InternVL3-78B.\cite{chen2024internvl}} to create challenging MCQA grounded in financial data visualizations. The process for generating each MCQA is as follows.

\begin{enumerate}
     \item \textbf{Temporal Context Sampling and Visualization:} For a given stock, relevant time intervals are sampled to define the temporal scope. Historical stock price data for the selected stock and period are then rendered as a time-series graph, providing  visual input.
    \item \textbf{Open-Ended Question and Answer Generation:} A VLM analyzes the generated stock price graph. Conditioned on the visual trends and the target temporal query type (Specific Time Period, Before/After, or Time Range) with few-shot instruction, it formulates an open-ended question and its corresponding answer. This step ensures the core question is directly derived from the underlying data pattern.
    \item \textbf{Intermediate Correctness Verification:} To mitigate potential VLM hallucinations or misinterpretations of the visual data, the generated open-ended question-answer pair undergoes a verification step. A separate VLM instance evaluates whether the answer aligns correctly with the provided graph and addresses the question. Pairs that fail this verification are discarded.
    \item \textbf{Structured Distractor Generation:} Given the generated open-ended question and its verified correct answer, distractors are formulated by LLM\footnote{The model used for distractor generation was LLaMA-3.3-70B-Instruct.}. This ensures that the distractors are textually coherent and challenge reasoning based on linguistic cues.  We employ few-shot prompting with the LLM, guiding it to generate distractors according to predefined strategies:(1) A choice describes a wrong trend that contradicts the answer; (2) A choice describes correct trend but associating it with an incorrect time period; (3) A choice is made up by the LLM.
    \item \textbf{Ensuring MCQA Correctness}
Ensuring the quality and unambiguous correctness of the generated MCQA is paramount for reliable evaluation. We adopt a correctness ensuring framework following VMCBench\cite{zhang2025automatedgenerationchallengingmultiplechoice}. Specifically, we employ a VLM\footnote{The VLM used for correctness ensure is InternVL3-78B.} to assess each generated MCQA. This VLM evaluates each distractor's similarity of to the correct answer and provides a 5-point Likert score reflecting the confidence that there is a single correct answer among the choices. A score of 5 indicates high confidence in the validity of question. We establish a threshold, discarding any generated MCQA receives a score less than 4.
\end{enumerate}
We recognized the limitations of fully automated assessments and added a human verification phase to improve the reliability of ADQAB. We manually reviewed a random sample of 20\% of the VLM-validated MCQA pairs. This review focused on checking the accuracy of the answers and ensuring clarity in the questions and choices. As a result, 94\% of the reviewed questions met our criteria for correctness and clarity, demonstrating the effectiveness of our VLM-based pre-filtering and the overall quality of the generated questions. The final evaluation set, generated from historical stock data, requires temporal reasoning and provides a reliable, challenging testbed for Time-Aware RAG systems.

\subsection{Benchmark Statistics}
\label{ssec:benchmark_statistics}
To provide a clear overview of the ADQAB components, Table \ref{tab:corpus_stats}, \ref{tab:eval_stats} summarizes the key statistics of the constructed corpus and the evaluation set.
\begin{table}[htbp]
\centering
\caption{ADQAB Corpus Statistics.}
\label{tab:corpus_stats}
\begin{tabular}{@{}lrr@{}}
\toprule
Article Type        & Count          & Avg. Words per Article \\
\midrule
Real News ($C_{real}$) & 23,737 & 840                    \\
Synthetic News ($C_{synth}$) & 3,300          & 192                   \\
\midrule
\textbf{Total Corpus ($C_{final}$)} & \textbf{27,037} & \textbf{761}           \\
\bottomrule
\end{tabular}
\\ 
\footnotesize 
\end{table}
\begin{table}[htbp]
\centering
\caption{ADQAB Evaluation Set ($Q_{eval}$)) Statistics by Temporal Query Type and Generated Sub-Type.}
\label{tab:eval_stats} 
\begin{tabular}{@{}p{0.25\linewidth} p{0.35\linewidth} r@{}}
\toprule
\textbf{Fundamental Query Type} & \textbf{Specific Generated Sub-Type} & \textbf{No. of Questions} \\
\midrule
Specific Time Period& Specific Year Trend & 75 \\
\midrule
\multirow{4}{*}{Before/After} & Before Year Anchor & 75 \\
                                     & Before Month Anchor & 75 \\
                                     & After Year Anchor & 75 \\
                                     & After Month Anchor & 75 \\
\midrule
\multirow{2}{*}{Time Range}   & Time Interval (Years) & 75 \\
                                     & Time Interval (Months) & 75 \\
\midrule
\multicolumn{2}{@{}l}{\textbf{Total Questions}} & \textbf{525} \\
\bottomrule
\end{tabular}
\end{table}
\begin{table*}[ht!]
\centering
\begin{tabular}{lccccc}
\toprule
\textbf{Method} & \textbf{Acc @ $k=5$} & \textbf{Acc @ k=10} & \textbf{Acc @ k=20} & \textbf{Acc @ k=50} \\
\midrule
BM25    &  36.72\% ($\pm$0.87\%) & 55.09\% ($\pm$0.58\%) & 65.79\% ($\pm$1.25\%) & 79.39\% ($\pm$0.69\%) \\
Naive RAG & 44.27\% ($\pm$0.67\%) & 56.72\% ($\pm$0.69\%) & 67.50\% ($\pm$1.19\%) & 74.82\% ($\pm$0.64\%) \\
Naive RAG + Reranker & 48.30\% ($\pm$0.50\%) & 59.39\% ($\pm$0.87\%) & 72.42\% ($\pm$0.53\%) & 82.10\% ($\pm$0.77\%) \\
TS-Retriever & 32.04\% ($\pm$0.25\%) & 42.78\% ($\pm$0.61\%) & 52.15\% ($\pm$0.75\%) & 58.51\% ($\pm$0.47\%) \\
TA-RAG + Reranker &  59.89\% ($\pm$0.98\%) & 71.50\% ($\pm$0.86\%) & 80.53\% ($\pm$0.53\%) & 87.09\% ($\pm$0.69\%)  \\
\textbf{TA-RAG (Full)} & \textbf{71.73\% ($\pm$0.81\%)} & \textbf{84.84\% ($\pm$0.46\%)}  & \textbf{88.23\% ($\pm$0.89\%)} & \textbf{88.00\% ($\pm$0.88\%)} \\
\bottomrule
\end{tabular}
\caption{Main results on the ADQAB. Mean accuracy (\%) with standard deviation ($\pm$\%) is reported for different retrieval methods and top-k settings over 5 runs. Best results are highlighted in \textbf{bold}.}
\label{tab:main_results_updated}
\end{table*}

\section{Experiments}
\subsection{Experimental Settings}
\textbf{Dataset}
All experiments were conducted using the ADQAB detailed in Section~\ref{sec:dataset}. \\
\textbf{Compared Methods.}
We compare our proposed TA-RAG approach with the following methods:
\begin{itemize}
    \item \textbf{BM25}\cite{robertson2009BM25}: BM25 is a widely used and traditional baseline method for information retrieval, which assigns a relevance score to documents based on the statistical relationship between the document content and the query terms.
    \item \textbf{Naive RAG} \cite{gao2024retrievalaugmented}: A standard baseline that retrieves chunks based on semantic similarity to the entire query, without explicit handling of temporal constraints.
    \item \textbf{Naive RAG + Reranker}: An extension of Naive RAG employing a two-stage retrieval process. It first retrieves a larger set of candidate chunks ($k \times 20$) using the Naive RAG approach. A subsequent reranking step then refines this set using a reranker model to select the final top $k$ most relevant chunks.
    \item \textbf{TS-Retriever} \cite{wu2024timesensitive}:  A novel training method integrates supervised contrastive learning emphasizing temporal constraints of the retriever.
    \item \textbf{TA-RAG + Reranker}: Following the same two-stage reranking process as Naive RAG + Reranker.
    \item \textbf{TA-RAG (Full)}: Our proposed framework (Section~\ref{TA-RAG}) incorporating dedicated components for temporal awareness throughout the RAG pipeline.
\end{itemize}
\textbf{Evaluation Metrics.}
Performance was measured using \textbf{Accuracy (Acc)}, calculating the percentage of MCQA questions where the model selected the correct answer choice. To ensure robust and reliable results, each experiment was conducted five times. The mean accuracy and standard deviation are reported.\\
\textbf{RAG pipeline components}
We implemented the RAG pipeline using LLaMA3.1-7B-Instruct for time information extraction and LLaMA3.3-70B-Instruct for question processing and answer generation. Document chunks and queries were embedded using the state-of-the-art "nomic-ai/nomic-embed-text-v1.5" model \cite{nussbaum2024nomic}, and where applicable, results were reranked using the "bge-reranker-v2-m3" model (568M parameters) \cite{chen2024bge}. Documents were split into non-overlapping chunks of up to 2048 tokens using SpaCy \cite{spacy2}, resulting in 77,965 chunks from 27,037 documents. Faiss\cite{johnson2019billion} and PyRange \cite{10.1093/bioinformatics/btz615} and used in the pipeline for data management.  

\textbf{Evaluation Protocol:} For all RAG methods evaluated on the MCQA task, the model first receives the question to perform the retrieval process. Subsequently, the generator LLM receives the retrieved context chunks along with the question stem and the multiple-choice options to select the final answer. This ensures the retrieval step operates solely based on the question, without access to the answer options. 
\begin{table*}[ht!]
    \centering
    \begin{tabular}{@{}lcccc@{}}
        \toprule
        Method                      & \textbf{Acc @ k=5}          & \textbf{Acc @ k=10}         & \textbf{Acc @ k=20}         & \textbf{Acc @ k=50}         \\ \midrule
        TA-RAG using $q_{core}$       & 67.05\% ($\pm$0.36\%)          & 80.50\% ($\pm$0.32\%)          & 88.15\% ($\pm$0.91\%)          & 88.19\% ($\pm$0.90\%)          \\
        TA-RAG w/o Time Filtering   & 67.16\% ($\pm$1.15\%)          & 80.88\% ($\pm$0.79\%)          & 87.09\% ($\pm$1.30\%)          & \textbf{88.65\% ($\pm$0.39\%)} \\
        TA-RAG w/o Context Structuring & 65.68\% ($\pm$0.95\%)          & 80.99\% ($\pm$0.78\%)          & 86.40\% ($\pm$0.32\%)          & 88.46\% ($\pm$0.77\%)          \\ \midrule
        \textbf{TA-RAG (Full)}      & \textbf{71.73\% ($\pm$0.81\%)} & \textbf{84.84\% ($\pm$0.46\%)} & \textbf{88.23\% ($\pm$0.89\%)} & 88.00\% ($\pm$0.88\%)          \\ \bottomrule
    \end{tabular}
    \caption{Ablation study results on ADQAB MCQA task. Mean Accuracy (\%) with standard deviation (over 5 runs) is reported. Best results are bolded.}
    \label{tab:ablation}
\end{table*}
\subsection{Main Result} 
To evaluate the effectiveness of our proposed TA-RAG framework, we conducted experiments on the ADQAB. The detailed results are presented in Table\ref{tab:main_results_updated}.

Notably, TA-RAG  achieves a mean accuracy of 71.73\% with only $k=5$ retrieved chunks, already surpassing other methods.  Its performance escalates with increasing k, reaching 84.84\% at $k=10$ and peaking at 88.23\% accuracy with $k=20$ chunks. This represents a substantial improvement over the Naive RAG baseline, for instance, yielding an accuracy gain of over 27\% at $k=5$ and 20.73\% at $k=20$. This robust performance underscores the efficacy of TA-RAG's temporally-aware retrieval in homing in on relevant evidence quickly. We observed the improvement of accuracy of TA-RAG converge sharply during $k$ increase, and there is a minimal decrease in accuracy to 88.00\% at $k=50$. To study this phenomenon, we selected one of the results within 5 experiment runs in which TA-RAG got 59 questions wrong when $k=20$ and 61 questions wrong when $k=50$, and only 31 of the questions were the same. We posit that this slightly declined with a large context ($k=50$) is because the very large context sizes begin to challenge the LLM's ability to discern the significant information, and encounters the "Lost-in-the-Middle" problem\cite{liu2023lostmiddlelanguagemodels}.

From the table, we observed that although TS-Retriever is designed for time-sensitive tasks, showed limited effectiveness on ADQAB. It is because current finetuning approach enhance retriever time constraint  with contrastive learning but no handling for Temporal Coverage and Temporal Endpoint Bias. This outcome shows that current specialized time-sensitive retrievers may struggle with the demands of complex diachronic questions, unlike the integrated temporal approach of TA-RAG.


The impact of incorporating reranking process yielded particularly insightful results regarding TA-RAG's architecture. Adding a general-purpose semantic reranker (TA-RAG + Reranker) consistently degraded TA-RAG's performance in all $k$ values; for example, accuracy dropped by approximately $13.34\%$ at $k=10$. This indicates that TA-RAG's initial retrieval mechanism is already highly attuned to temporal pertinence. A standard semantic reranker, lacking this specialized temporal awareness, may inadvertently disrupt this by deprioritizing documents that are temporally crucial yet perhaps not maximally semantically similar by the reranker's general criteria. While such rerankers can provide modest gains for temporally naive methods like Naive RAG, they are counterproductive for TA-RAG. This finding strongly reinforces the efficacy of an end-to-end, temporally-aware retrieval strategy as realized in TA-RAG.

While TA-RAG achieves strong performance, we acknowledge several limitations. The offline time information extraction incurs computational costs that scale with corpus size, though it does not affect online latency. Both time extraction and question processing depend on the temporal reasoning capabilities of LLMs. Additionally, the extra LLM call for question processing increases execution time by approximately 1.4× compared to standard Naive RAG. These findings highlight clear opportunities for future optimization.
\subsection{Ablation Study}
To assess the individual contributions of TA-RAG's core modules, we conducted an ablation study, with results presented in Table~\ref{tab:ablation}. We systematically evaluated the impact of removing Hypothetical Temporal Query Embeddings ($e_{hypo}$), Time Filtering, and Temporal Context Structuring.

\textbf{Impact of Hypothetical Temporal Query Embeddings ($e_{hypo}$).}
The $e_{hypo}$ component is designed to enrich query representations with temporal nuances, thereby enhancing retrieval precision. When this component was removed ("TA-RAG using $q_{core}$"), we observed notable performance declines for $k \le 20$: accuracy dropped by $4.68\%$ at $k=5$ (from $71.73\%$ to $67.05\%$), $4.34\%$ at $k=10$, and a small difference $0.08\%$ at $k=20$. These results underscore the value of $e_{hypo}$ in generating effective query representations, particularly when the number of retrieved documents is limited and precision is paramount.

\textbf{Impact of Time Filtering.}
The Time Filtering stage aims to proactively narrow the search space to documents that are temporally congruent with the query. Disabling this module ("TA-RAG w/o Time Filtering") consistently reduced accuracy for retrieval depths up to $k=20$. Specifically, performance fell by $4.57\%$ at $k=5$, $3.96\%$ at $k=10$, and $1.14\%$ at $k=20$. This confirms the efficacy of Time Filtering in focusing the retrieval on temporally relevant candidates, which is especially beneficial in scenarios emphasizing retrieval precision with fewer items.

\textbf{Impact of Temporal Context Structuring.}
Temporal Context Structuring is responsible for organizing the retrieved chunks in a manner that aids the generator LLM's synthesis of temporally complex information. Its removal ("TA-RAG w/o Context Structuring") led to the most substantial performance degradation at $k=5$, with accuracy decreasing by $6.05\%$ (from $71.73\%$ to $65.68\%$). The negative impact persisted with drops of $3.85\%$ at $k=10$ and $1.83\%$ at $k=20$. This highlights the importance of a structured contextual input for the LLM, particularly when it processes a smaller, more concentrated set of temporal evidence.

A different trend emerged when the retrieval was significantly increased to $k=50$. In this scenario, the full TA-RAG framework ($88.00\%$) was marginally outperformed by the three ablated variants. Specifically, "TA-RAG using $q_{core}$" achieved $88.19\%$ (+$0.19\%$ difference), "TA-RAG w/o Time Filtering" reached $88.65\%$ (+$0.65\%$ difference and the highest accuracy at this $k$), and "TA-RAG w/o Context Structuring" scored $88.46\%$ (+$0.46\%$ difference). These differences are relatively small and, may not all indicate significant that the full TA-RAG model perform poorly. Given the size of the test set (525 questions), these subtle changes may stem from the complex interaction of high recall, the challenges LLMs encounter with very extensive contexts, such as the "Lost-in-the-Middle" effect\cite{liu2023lostmiddlelanguagemodels}. 

In summary, our ablation study confirms that Hypothetical Temporal Query Embeddings($e_{hypo}$), Time Filtering, and Temporal Context Structuring are all valuable components of TA-RAG, contributing significantly to its performance, especially for retrieval depths with smaller $k$. The nuanced results at $k=50$ underscore the challenges of processing very large contexts with current LLMs. These findings highlight the general robustness of our integrated approach, while also suggesting avenues for future research into strategies tailored for varying retrieval depths and context sizes.


\section{Conclusion}
We proposed TA-RAG, a framework incorporating explicit temporal processing across the retrieval and generation pipeline, to address the limitations of standard RAG systems for time-sensitive and diachronic queries. We also presented ADQAB, a new benchmark specifically designed for evaluating complex temporal question answering. Experiments demonstrate TA-RAG's effectiveness in handling temporal constraints compared to baselines. While promising, the reliance on LLMs and observed challenges with standard retrieval and ranking methods highlight the need for future research into more efficient and robust techniques for temporal information representation, retrieval, and reasoning within augmented generation systems.

\bibliographystyle{ACM-Reference-Format}
\bibliography{reference}

\end{document}